\documentclass{article}

\usepackage{arxiv}

\usepackage[utf8]{inputenc} 
\usepackage[T1]{fontenc}    
\usepackage{hyperref}       
\usepackage{url}            
\usepackage{booktabs}       
\usepackage{amsfonts}       
\usepackage{nicefrac}       
\usepackage{microtype}      
\usepackage{lipsum}
\usepackage{graphicx} 
\usepackage{amsthm}
\usepackage{amssymb}
\usepackage{amsmath}
\usepackage{todonotes} 
\usepackage{multirow}
\usepackage[ruled]{algorithm}
\usepackage{algorithmic}
\newcommand{\argmax}[1]{\underset{#1}{\operatorname{arg}\,\operatorname{max}}\;}

\title{Deep Submodular Networks for Extractive Data Summarization}

\author{
  Suraj Kothawade\\
  University of Texas, Dallas\\
  \texttt{suraj.kothawade@utdallas.edu} \\
   \And
 Jiten Girdhar \\
 University of Texas, Dallas\\
 \texttt{jiten.girdhar@utdallas.edu} \\
  \And
 Chandrashekhar Lavania \\
 University of Washington, Seattle\\
 \texttt{lavaniac@uw.edu} \\
  \And
 Rishabh Iyer\\
 University of Texas, Dallas\\
 \texttt{rishabh.iyer@utdallas.edu} \\
}

\begin{document}
\maketitle

\begin{abstract}
Deep Models are increasingly becoming prevalent in summarization problems (e.g. document, video and images) due to their ability to learn complex feature interactions and representations. However, they do not model characteristics such as diversity, representation, and coverage, which are also very important for summarization tasks. On the other hand, submodular functions naturally model these characteristics because of their diminishing returns property. Most approaches for modelling and learning submodular functions rely on very simple models, such as weighted mixtures of submodular functions. Unfortunately, these models only learn the relative importance of the different submodular functions (such as diversity, representation or importance), but cannot learn more complex feature representations, which are often required for state-of-the-art performance. We propose \emph{Deep Submodular Networks} (DSN), an end-to-end learning framework that facilitates the learning of more complex features and richer functions,  crafted for better modelling of all aspects of summarization. The DSN framework can be used to learn features appropriate for summarization from scratch. We demonstrate the utility of DSNs on both generic and query focused image-collection summarization, and show significant improvement over the state-of-the-art. In particular, we show that DSNs outperform simple mixture models using off the shelf features. Secondly, we also show that just using four submodular functions in a DSN with end-to-end learning performs comparably to the state-of-the-art mixture model with a hand-crafted set of 594 components~\cite{tschiatschek2014learning} and outperforms other methods \cite{tschiatschek2018differentiable, dolhansky2016deep} for image collection summarization.
\end{abstract}


\section{Introduction} \label{sec::Intro}


The volume of data generated daily has seen exponential growth in recent years \cite{patrizio2018idc}, coming from a number of sources including video streams, images, sensor streams, and text data. However, this large data volume is often redundant, and processing it is a non-trivial and expensive task. Extracting meaningful information through summarization can make it easier to process by downstream tasks. Extractive summarization can be viewed as a constrained subset selection problem. 

Submodular functions are a natural choice for summarization due to their diminishing returns property, since they naturally model aspects like diversity, coverage and representation. As a result, they have been used in various data summarization problems like image collection summarization \cite{tschiatschek2014learning, simon2007scene}, video summarization ~\cite{zhang2016video,Gygli2015VideoSB,Kaushal2019DemystifyingMV,Kaushal2019AFT, kaushal2019framework, kaushal2020realistic}, document summarization ~\cite{lin2011class,lin2012learning,li2012multi,chali2017towards,yao2017recent}, and data selection~\cite{wei2015mixed,wei2015submodularity,wei2014submodular,kaushal2019learning}. Most work on learning submodular models has either focused on very simple models such as mixtures of submodular functions~\cite{lin2012learning,tschiatschek2014learning,gygli2015video}, or are rich but restricted to sub-classes of functions (like log-determinants~\cite{kulesza2012determinantal} or nested concave over submodular functions~\cite{dolhansky2016deep}). On the other hand, with the success of deep learning based methods, deep features have replaced many older techniques using hand-crafted features due to their ability in capturing semantic information through feature interactions. This has led to the use of deep models in various summarization problems \cite{fu2019attentive, zhang2016video}. Deep models however, do not model interaction between instances required for modelling diversity, representation and coverage.  In this paper, we combine the best of both worlds by jointly learning submodularity and feature interactions leading to richer and deeper summarization models. 

 Formally, a submodular function is defined as follows. A set function $f: 2^V \rightarrow \mathbf{R} $, defined over all subsets of a ground set $V = \{v_1,v_2, \cdots ,v_n\}$ (images, sentences, or video frames), is \textit{submodular} if for $x \in V$, $f(A \cup x) - f(A )\geq f(B \cup x) - f(B)$, $\forall A \subseteq B \subseteq V$ and $x \notin B$. For a set $A \subseteq V$, $f(A)$ provides a real-valued score for A. In the context of summarization, $f(A)$ indicates the quality of summary defined by set A. A function $f$ is said to be monotone if $f(A) \leq f(B)$ whenever $A \subseteq B$. Also, $f$ is \textit{supermodular} if $- f$ is \textit{submodular}, modular if it is both, and \textit{normalized} if
$f(\phi) = 0$. 

Any submodular function based summarization procedure consists of at least two parts: 1) Defining the objective submodular function and 2) performing a constrained maximization on the function. Even though constrained submodular maximization is NP-hard, there are a number of constant factor approximations available ~\cite{minoux1978accelerated, mirzasoleiman2014lazier, mirzasoleiman2016fast, nemhauser1978analysis}. However, the optimal design of the objective function for a particular problem is a non-trivial task, and even learning the submodular function from data is extremely hard and not easy to even approximate~\cite{balcan2011learning}. However, learning can be made easier by restricting the scope of the submodular function. In this work, we present a rich and powerful, yet tractable framework called Deep Submodular Networks (DSNs) which enables joint learning of features and function parameters in an end-to-end and tractable manner. Thus, learning using DSN amplifies the modelling power by allowing us to represent various complex submodular functions. Moreover, the DSN framework has the following characteristics: 1) can be applied to any summarization task by parameterizing submodular functions, 2) can be operated on multi-modal data if the representation can be done in the same feature space (as in query focused summarization) 3) allows warm-start by instantiating functions using features from a pre-trained network 4) reduces computational complexity by learning complex representation using fewer components that would otherwise need many components. We then study the parameter learning problem with DSNs, and propose a max-margin based learning algorithm. Finally, we present results on the task of summarizing an image collection. Two flavours of summarization are considered: 1) generic and 2) query-focused. The results demonstrate that mixture learning using DSN is effective.





\section{Related Work and Our Contributions}
 \subsection{Related work on Extractive Data Summarization: } The problems of image, video, and document summarization have been extensively explored in literature. For image collection summarization \cite{simon2007scene} used the facility location function with a diversity penalty term and \cite{sinha2011extractive} propose a diversity model by using a coverage function and a disparity function. Other approaches include hierarchical clustering \cite{jaffe2006generating} and k-medoids-based clustering~\cite{zhao2016visual}, dictionary learning~\cite{yang2013image}, and deep learning~\cite{ozkose2019diverse}. \cite{tschiatschek2014learning} was among the first to learn the weights of a mixture of submodular functions using max-margin learning. A related approach is the use of $k$-DPPs~\cite{Kulesza2011kDPPsFD}. \cite{kaushal2020unified} extends the dataset of \cite{tschiatschek2014learning} for query focused summarization, which we use for our experiments. Similar to image summarization, submodular optimization is vastly used for video summarization~\cite{zhang2016video,Gygli2015VideoSB,Kaushal2019DemystifyingMV,Kaushal2019AFT} and document summarization~\cite{lin2011class,lin2012learning,li2012multi,chali2017towards,yao2017recent}. Other diversity-based approaches for video summarization include using Maximum Marginal Relevance (MMR) \cite{li2012multi}, and Determinantal Point Processes (DPPs) for summarization~\cite{zhang2016video, gong2014diverse, kulesza2012determinantal}.

\subsection{Related work on Query Focused Data Summarization: } While most literature has focused on generic summarization, a recent work employs submodular functions to parametrize submodular mutual information measures for query-focused summarization \cite{kaushal2020unified}. \cite{lin2011class} achieves state-of-the-art results for query-focused document summarization by defining a joint diversity and query relevance term. In video summarization, \cite{vasudevan2017query} study a query relevance term based on graph-cut function and ~\cite{sharghi2016query,sharghi2017query} study hierarchical DPPs to model both diversity and query relevance. Graph-cut functions are also used by \cite{li2012multi} for query focused and update summarization. All the above work on query-focused summarization use instances of submodular information measures. In our work, we use a general parametrized form of submodular mutual information measures as proposed by \cite{kaushal2020unified}.

\subsection{Related work on Learning Submodular Functions: } Since submodular functions lie in a vast space with $2^n$ degrees of freedom, it is non-trivial to find the right submodular function for a given problem. Hence, it is imperative to learn submodular functions based on queries or data. However, many theoretical results show that learning submodularity can be very hard in the worst case -- for example, learning general submodular in the PMAC setting is not approximable to any polynomial factor in the worst case \cite{balcan2011learning}. The problem becomes easier if we restrict ourselves to simpler classes of functions. One example of this is by defining a simple mixture of submodular functions~\cite{tschiatschek2014learning, kaushal2019framework, kaushal2020realistic, sipos2012large, lin2012learning,lavania2019auto} by constructing a weighted convex combination of non-negative submodular functions $F_w(A) = \sum_{i=1}^M w_if_i$, where $f_1, f_2, \cdots ,f_M$ are submodular functions with weights $w = (w_1, w_2, \cdots, w_M)$, $w_i  \geq 0$, $\sum_i w_i = 1$. Here, the mixture's components $f_i$ are fixed (instantiated using features from a pre-trained network), and only the mixture weights are learned.  However, this limits the modelling power of $F_w(.)$ because the feature interactions and the functions' internal parameters are not learned. To mitigate this, recent work on a richer parametric family of submodular functions called deep submodular functions or DSFs \cite{dolhansky2016deep, bilmes2017deep} explores a new class of functions with characteristics similar to a DNN due to its multilayered architecture.  They try to address this by parameterizing $f_{\hat{w}} \in DSF$ where the vector $\hat{w}$ determines the network topology and the concave functions. DSFs employ a recursive application of concave over modular functions, thereby maintaining submodularity. 
Although DSFs boost the modelling power of submodular functions, they are still a restricted subclass of functions and do not effectively learn the feature representations in an end to end manner. As a result, DSFs often perform worse in practice compared to mixtures of many submodular functions (as we see in our experiments).  

\subsection{Our Contributions}
The main contributions of this work are as follows. We present Deep Submodular Networks (DSNs) which enables joint learning of the model parameters and feature representations. We then study the parameter learning problem for DSNs via max-margin learning and study the properties of the resulting algorithm. We also show how the DSNs for different submodular functions can be combined either via mixtures of submodular functions or via compositions of submodular functions, similar to deep submodular functions. We then empirically demonstrate the effectiveness of DSN for data summarization by drawing various insights. a) We first show that DSN can learn superior features that, when used to instantiate submodular functions, generate summaries with higher V-ROUGE scores compared to submodular functions instantiated using features from a pre-trained DNN. b) We next show that using a simple mixture of four DSNs significantly outperforms existing baselines including a mixture of four submodular functions (without learning the feature interactions), the deep submodular functions~\cite{dolhansky2016deep}, and differentiable submodular maximization~\cite{tschiatschek2018differentiable}. Furthermore, we also show that just using 4 DSN components, we are able to achieve comparable performance to a mixture model of 594 hand-crafted components. To the best of our knowledge, an end-to-end learning framework like DSN has not been studied before for any summarization task.

\section{Submodular Functions for Summarization}

We focus on monotone, non-negative submodular functions for use as components in the submodular mixture. These mixture components will belong to one of the following types of submodular functions, depending on the type of summarization task in hand. For a more detailed summary of submodular functions used, see~\cite{iyer2015submodular,iyer2015spp,kaushal2020unified,iyer2020submodular}.

\subsection{Generic Summarization}

In this paper, we focus on generic extractive summarization under a cardinality constraint, the optimization problem is:  $\max_{A: |A| \leq k}  f(A)$ for a given submodular function $f$ and budget $k$. Submodular functions for this flavour of summarization are mostly instantiated using either a similarity matrix $S$ that stores the similarity between data points, or directly using the feature representation of data points. Let $V$ be the ground set of data points which need to be summarized. We denote $A \subseteq V$ as a candidate subset of the data points in $V$. For any two data points $i,j \in V$, we denote $s_{ij}$ to be the similarity between them.\\
\noindent\textbf{Facility Location: } The facility location function can be defined as: 
\begin{align} \label{eq:FL}
f_{FL}(A) = \sum_{i \in V} \max_{j \in A} s_{ij}    
\end{align}
An additional penalty term $\lambda$ can be added to get:
\begin{align} \label{eq:FLP}
f_{FLP}(A) = \sum_{i \in V} \max_{j \in A} s_{ij} - \lambda \sum_{i,j \in A} s_{ij}
\end{align}

Instantiating kernel based functions like facility location function is computationally expensive as it requires the generation of the similarity values $s_{ij}$. The similarity calculation involved the computation of a $O(n^2)$ kernel matrix \cite{liu2013submodular}. Facility location function models representation, and is similar to k-mediod clustering and has been used in various summarization tasks \cite{simon2007scene, takamura2010text, xu2019deep}.

\noindent\textbf{Saturated Coverage: } This function models coverage of the ground set and is defined as: 
\begin{align} \label{eq:SatCov}
f_{SAT}(A) = min(\sum_{i \in A} s_{ij}, \lambda \sum_{i \in V} s_{ij})
\end{align}
where $\lambda$ value controls the degree at which this function gets saturated. Saturated Coverage allows the selection of multiple representatives of every category, whereas facility location prefers only a single representative.

\noindent\textbf{Graph Cut: } The generalized graph cut is defined as:
\begin{align} \label{eq:GraphCut}
f_{GC}(A) = \lambda \sum_{i \in V} \sum_{j \in A} s_{ij} - \sum_{i,j \in A} s_{i,j}
\end{align}
This function is similar to saturated coverage and facility location in terms of its modelling behavior. The $\lambda$ parameter tunes the tradeoff between diversity and representation. As $\lambda$ increases, graph-cut tries to model diversity. For values of $\lambda < 0.5$ graph-cut is monotone submodular and for $\lambda > 0.5$ it is non-monotone submodular. It has been widely used for summarization tasks \cite{vasudevan2017query, li2012multi, lin2012submodularity}.

\noindent\textbf{Feature Based: } The feature based function assumes that each data element from the ground set $V$ can be represented in a feature space $U$. They have proved to be useful in a variety of summarization tasks ~\cite{ni2015submodular, wei2014submodular, tschiatschek2014learning, kirchhoff2014submodularity}. This function can be defined as:

\begin{align} \label{eq:FB}
f_{FB}(A) = \sum_{i \in U} \gamma_i\psi_i(m_i(A))
\end{align}

Here $m_i(A) = \sum_{a \in A} U_i^a$ with $U_i^a$ being the representation of data element $a$ for feature $i$. The functions $\psi_{i}(\cdot)$ are monotone non-decreasing concave functions associated with feature $i$ and $\gamma_i$ is the corresponding weight. The maximization of a feature based function attempts to find a subset $A$ that covers the maximum number of concepts in $V$.


\subsection{Query-focused Summarization}
Since generic summarization focuses on creating a general candidate summary by treating all data points in the ground set $V$ equally, it is impossible to incorporate user intent within a summary. Query focused summarization attempts to produce summaries relevant to the input query set $Q$ in addition to the characteristics of a good summary (such as representation, coverage, diversity). In this paper, we use the recently proposed framework of~\cite{iyer2020submodular,kaushal2020unified} which uses the submodular mutual information (SMI). SMI is a generalization to handle query relevance in a natural manner. Given a set $B \subseteq (V \cup Q)$, the submodular mutual information $I_f(A; B)$ is defined as: $I_f(A; B) = f(A) + f(B) - f(A \cup B)$. We define the optimization problem for cardinality constrained query-focused summarization to be $\max_{A \subseteq V: |A| \leq k}  I_f(A; Q)$ for a given submodular function $f: 2^{V \cup Q} \rightarrow \mathbf{R}$, query set $Q$ and budget $k$. In such a problem formulation, although $f$ is defined on $V \cup Q$, only items from $V$ are summarized, i.e, the discrete optimization will only be on subsets of $V$.
In addition to the similarity matrix $S$, we compute $S_Q$, which stores the similarity between each data point of $V$ with every query data point in $Q$. For any two data points $i \in V$ and $j \in Q$, we denote $sq_{ij}$ to be the similarity between them.

\noindent\textbf{Graph Cut Mutual Information:} The submodular mutual information (SMI) instantiation of graph-cut (GCMI) is defined as:
\begin{align} \label{eq:GCMI}
I_f(A;Q)=2\sum_{i \in A} \sum_{j \in Q} sq_{ij}
\end{align}
Since maximizing GCMI maximizes the joint pairwise sum with the query set, it will lead to a summary similar to the query set $Q$. In fact, specific instantiations of GCMI have been intuitively used for query-focused summarization for videos ~\cite{vasudevan2017query} and documents ~\cite{lin2012submodularity, li2012multi}. 


\noindent\textbf{Facility Location Mutual Information - V1:} In the first variant of FL, we set $D$ to be $V$. The SMI instantiation of FL1MI can be defined as:
\begin{align} \label{eq:FL1MI}
I_f(A;Q)=\sum_{i \in V}\min(\max_{j \in A}s_{ij}, \lambda \max_{j \in Q}sq_{ij})
\end{align}
The first term in the min(.) of FL1MI models diversity, and the second term models query relevance. An increase in the value of $\lambda$ causes the resulting summary to become more relevant to the query.

\noindent\textbf{Facility Location Mutual Information - V2:} In the V2 variant, we set $D$ to be $V \cup Q$. The SMI instantiation of FL2MI can be defined as:
\begin{align} \label{eq:FL2MI}
I_f(A;Q)=\sum_{i \in Q} \max_{j \in A} sq_{ij} + \lambda\sum_{i \in A} \max_{j \in Q} sq_{ij}
\end{align}
FL2MI is very intuitive for query relevance as well. It measures the representation of data points that are the most relevant to the query set and vice versa. It can also be thought of as a bidirectional representation score. 

\section{Deep Submodular Networks}
This section presents an efficient learning framework to learn a mixture of submodular functions in an end-to-end manner. Learning mixtures of submodular functions has given good empirical results on various summarization tasks including image \cite{tschiatschek2014learning}, video \cite{kaushal2020realistic} and document  summarization \cite{lin2012learning}. They use a max-margin learning framework for effectively learning the weights of the mixture. The weights are optimized such that the learned function tries to mimic characteristics of ground truth summaries (typically human-generated) based on a given loss. This is done by computing a gradient that requires minimizing the difference between two submodular components (one is the mixture and other is the loss) \cite{tschiatschek2014learning}. In our work, we apply the max-margin approach for learning the parameters of DSN end-to-end. We now describe the architecture and learning of DSNs.

\subsection{Architecture} \label{sec::DSNArch}

\begin{figure}[h]
    \centering
    \includegraphics[width=0.8\textwidth]{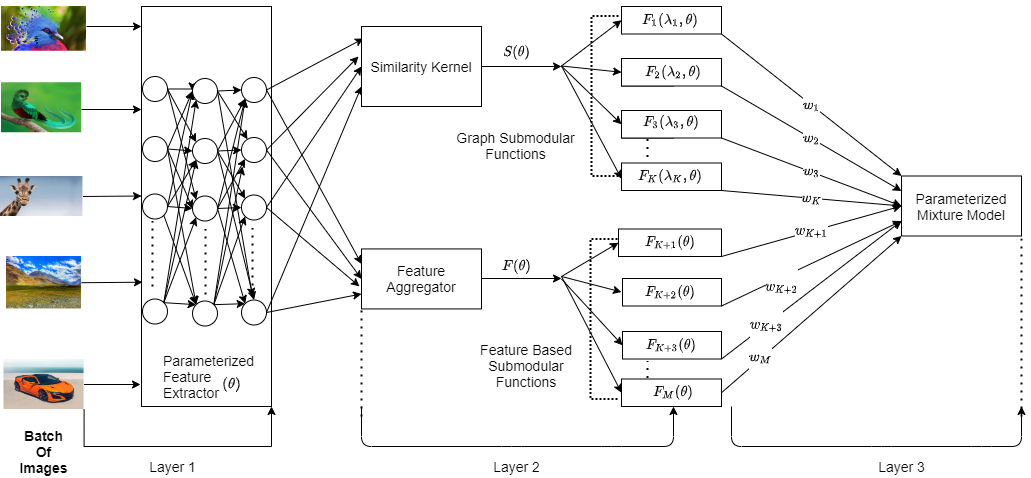}
    \caption{Deep Submodular Networks: A three layer architecture. First is the feature extractor module, which is a parameterized neural network. Second is the similarity kernel and feature aggregator modules to initlialize either the similarity based or feature based submodular functions with internal parameters $\lambda_i$.} 
    \label{fig:dsn}
\end{figure}

Figure~\ref{fig:dsn} describes the three-layer architecture of DSN. In the first layer, given a set of data points, which can be either be represented by using features from a pre-trained DNN or by raw features, we compute its DSN feature representation by passing it through a feature extractor which is parameterized by $\theta$. Here,  $\theta$ is a common parameter for all the DSN components and is responsible for learning feature interactions. As discussed above, submodular functions can be instantiated by using features (as in feature based functions) or by a similarity kernel (as in graph submodular functions). In addition to $\theta$, the submodular functions are parameterized using their internal parameters $\lambda_i$ in the second layer of DSN. For instance, $\lambda_i$ can be the diversity-representation tradeoff parameter in graph-cut or the saturation weight in saturated coverage or the query-relevance tradeoff in FLMI functions. The last layer is a convex combination of these submodular functions parameterized by $\theta$ and $\lambda$, where the mixture weights are parameterized by $w_i$. For generic summarization, we define the DSN mixture model as :
\begin{align} \label{mixture-dsn-gen}
    F(A, x, w, \theta, \lambda) = \sum_{i=1}^M w_i f_i(A, x, \theta, \lambda_i)
\end{align}
Here $x$ is the raw features of the input instances (for e.g. a video or image collection). For query-focused summarization, a Query set $Q$ can be incorporated in this model by defining a DSN over the submodular mutual information based functions:
\small{
\begin{align} \label{mixture-dsn-query}
    F(A, Q, x, w, \theta, \lambda) = \sum_{i=1}^M w_i I_{f_i}(A; Q, x, \theta, \lambda_i)
\end{align}}
\normalsize
The DSN optimization problem for summarization is: 
\begin{align}
A \in \argmax{A \subseteq V, |A| \leq k} F(A)    
\end{align}
where $F$ is defined in equations~\eqref{mixture-dsn-gen} and \eqref{mixture-dsn-query}.

Finally, we can also consider more complex ways of combining the DSNs as opposed to just simple mixtures. In particular, since concave over monotone submodular functions are submodular~\cite{dolhansky2016deep}, we can combine them recursively using an approach very similar to a DSF. Given concave function $\psi_i, i = 1, 2, \cdots, P$, define the following parameterized model:
\begin{align} \label{dsf-dsn-gen}
    F(A, x, w, \theta, \lambda) = \sum_{i = 1}^P \psi_i(\sum_{j=1}^M w_{ij} f_i(A, x, \theta, \lambda_i))
\end{align}
We can similarly define this for submodular mutual information functions (i.e. equation~\eqref{mixture-dsn-query}) for query based summarization. For completeness, we add it below.
\begin{align} \label{dsf-dsn-query}
    F(A,Q, x, w, \theta, \lambda) = \sum_{i = 1}^P \psi_i(\sum_{j=1}^M w_{ij} I_{f_i}(A;Q,x, \theta, \lambda_i))
\end{align}

Finally, note that we can also recursively define to have additional compositions of concave functions. In the rest of the paper, we will derive expressions for simple mixtures (i.e. equations~\eqref{mixture-dsn-gen} and \eqref{mixture-dsn-query}), but the expressions for the more general compositions (i.e. DSF-DSNs) can be obtained similarly.

\subsection{Learning Deep Submodular Networks} \label{sec::MaxMarginDSN}

In this paper, we consider the supervised summarization problem. In the case of generic (i.e. query agnostic) summarization, we assume we are given a dataset of $N$ training examples as $(Y^{(n)}, V^{(n)}, x^{(n)})$ where $n=1 \dots N$, $Y^{(n)}$ is a human summary for the $n^{th}$ ground set $V^{(n)}$ with off-the-shelf/raw features $x^{(n)}$. Also assume, we have an ideal loss function (equivalently evaluation function) $l : 2^V \times 2^V \rightarrow \mathbf{R}$
which measures how close an automatically generated summary is to a human summary. The goal of the learning problem is to learn parameters $\theta, w, \lambda)$ which solves the following empirical risk minimization problem:
\begin{align}\label{empriskmin}
    \min_{w, \theta, \lambda} \sum_{n = 1}^N l(\argmax{A \in C^{(n)}} F(A, x^{(n)}, w, \theta, \lambda), Y^{(n)})
\end{align}
Here the constraint $C^{(n)} = \{A \subseteq V^{(n)}, |A| \leq k\}$. 
In the case of query based summarization, the setup is very similar~\cite{kaushal2020unified}. The training data here additionally has queries for every example: $(Q^{(n)}, Y^{(n)}, V^{(n)}, x^{(n)})$ where $n=1 \dots N$. The resulting learning problem is very similar to equation~\eqref{empriskmin} except that we use the query based model and therefore have the query as well. 

While this is the \emph{true} learning problem we care about, it is very difficult to optimize this since this is a non-convex and has sharp discontinuities. As a result, we resort to max-margin based learning framework.

Recall that in our model, $\lambda_i$ corresponds to the internal parameters of the submodular function $i$, and define the vector $\lambda = (\lambda_1, \dots, \lambda_M)$. Similarly,  $\theta$ represents the shared parameters of the learned features, and $w$ denotes the weight matrix (which is a simple vector in the mixture model case and a matrix in the DSF case). Let $\Theta = (w,\lambda, \theta)$ denote the vector of parameters of the DSN. The max-margin approach \cite{taskar2004max, sipos2012large, lin2012learning, tschiatschek2014learning, kaushal2020realistic} is used to  learn the parameters $\Theta$ of $F(A, x, \Theta)$. The goal of max-margin is to try to ensure that we learn parameters $\Theta$ such that the following inequality holds for every training example: 
\small{
\begin{align}
    F(Y^{(n)}, x^{(n)}, \Theta) \geq \max_{A \in C^{(n)}}[F(A, x^{(n)}, \Theta) + l(A, Y^{(n)})]
\end{align}}
\normalsize
Following~\cite{taskar2004max, tschiatschek2014learning,lin2012learning}, we can formulate this as a loss-augmented inference based optimization problem. Define:
\begin{multline}
L_n(\Theta) = (\max_{A \in C^{(n)}} [\sum_{i=1}^M w_i f_i(A, x^{(n)}, \theta, \lambda_i) + l_{Y^{(n)}}(A)] - \sum_{i=1}^M w_i f_i(Y^{(n)}, x^{(n)}, \theta, \lambda_i)) +\frac{\beta}{2} ||w||_2^2 
\label{mmleq}
\end{multline}
where, $\beta$ is the regularization parameter. Then the learning problem is: $\min_{\Theta} \sum_{n = 1}^N L_n(\Theta)$.

For the purpose of learning the parameters $w_i$ , $\lambda_i$ and $\theta$ we compute the subgradients for each of the parameters as follows,\\ 
\textbf{Sub gradient of mixture weights $w_i$:}
\begin{align}
\frac{\partial L_n}{\partial w_i} = f_i(\hat{A}_n, x^{(n)}, \theta, \lambda_i) - f_i(Y^{(n)}, x^{(n)}, \theta, \lambda_i) + \beta_iw_i
\label{eq2}
\end{align}
\textbf{Sub gradient of function parameters $\lambda_i$:}
\begin{align}
\frac{\partial L_n}{\partial \lambda_i} = w_i \frac{\partial f_i(\hat{A}_n, x^{(n)}, \theta, \lambda_i)}{\partial \lambda_i}  - w_i \frac{\partial f_i(Y^{(n)}, x^{(n)}, \theta, \lambda_i)}{\partial \lambda_i}
\label{eq3}
\end{align}
\textbf{Sub gradient of common feature parameters $\theta$:}
\begin{align}
\frac{\partial L_n}{\partial \theta} = w_i \frac{\partial f_i(\hat{A}_n, x^{(n)}, \theta, \lambda_i)}{\partial \theta}  - w_i \frac{\partial f_i(Y^{(n)}, x^{(n)}, \theta, \lambda_i)}{\partial \theta}
\label{eq4}
\end{align}


where, $\hat{A}_n = \argmax{A \in C^{(n)}} F(A, x^{(n)}, \theta, w, \lambda) + l_{Y^{(n)}}(A) $

We summarize our learning methodology in algorithm \ref{alg:dsnlearn}.

\subsection{Subgradients of submodular functions}

\subsubsection{Generic Summarization}
Consider an entry $s_{ij}$ from the similarity matrix $S$ which is computed from DSN feature representations $X_i=\sigma(\theta^T \cdot U_i)$ and $X_j= \sigma(\theta^T \cdot U_j)  $, where $U_i$ and $U_j$ represent off-the-shelf/raw features of data point $i$ and $j$ respectively. Let us assume that that $X_i$ and $X_j$ are L2 normalized. The cosine similarity between them is simply their dot product: $s_{ij} = X_i^T \cdot X_j$. We can rewrite $s_{ij} = \sigma(\theta^T \cdot U_i)^T \cdot \sigma(\theta^T \cdot U_j)$, where $\sigma$ is the sigmoid function. The subgradient of the similarity matrix $\triangledown_\theta s_{ij}$ can be computed as follows:


\begin{align}\label{eq: gradCalcFL4}
\triangledown_\theta s_{ij} = \sigma(\theta^T U_i)^T\sigma(\theta^T U_j)((1-\sigma(\theta^TU_j))^T U_j + (1- \sigma(\theta^T U_i ))^T U_i)
\end{align}

We compute $\triangledown_{\lambda}f$, the subgradient with respect to the respective internal parameters of individual function components, and  $\triangledown_\theta s_{ij}$, the subgradient with respect to parameters of learned features. These subgradients are computed for each of the functions used in the mixture as follows:

\noindent\textbf{Facility Location with Penalty:} From \ref{eq:FLP}, we get:
\begin{equation}
\begin{aligned}
\triangledown_\theta f_{FLP} &= \sum_{i \in V}\triangledown_\theta s_{ij_i} - \lambda \sum_{i,j \in S} \triangledown_\theta s_{ij}\\
\triangledown_\lambda f_{FLP} &= -\sum_{i,j \in S} s_{ij}
\end{aligned}
\end{equation}

\noindent where, $j_i = \argmax{j \in S}s_{ij}$\\

\noindent\textbf{Saturate coverage:} From \ref{eq:SatCov}, we get:
\begin{equation}
\begin{aligned}
\triangledown_\lambda f_{SC} &= \sum_{i,j \in V}s_{ij}*1_{\lambda \sum_{j \in V} s_{ij} \leq \sum_{j \in A} s_{ij}}\\
\triangledown_\theta f_{SC} &= \sum_{i \in V}(\sum_{j \in A} \triangledown_\theta s_{ij}*1_{ \sum_{j \in A} s_{ij} \leq \lambda \sum_{j \in V} s_{ij}} + \lambda \sum_{j \in V} \triangledown_\theta s_{ij}*1_{\lambda \sum_{j \in V} s_{ij} \leq \sum_{j \in A} s_{ij}})
\end{aligned}
\end{equation}

\noindent\textbf{Graph Cut:} From \ref{eq:GraphCut} we get:
\begin{equation}
\begin{aligned}
\triangledown_\lambda f_{GC} &= \sum_{i \in V} \sum_{j \in A} \triangledown_\theta s_{ij}\\
\triangledown_\theta f_{GC} &= \lambda \sum_{i \in V} \sum_{j \in A} \triangledown_\theta s_{ij} - \sum_{i,j \in A} \triangledown_\theta  s_{i,j}
\end{aligned}
\end{equation}

\noindent\textbf{Feature based:} From \ref{eq:FB} and assuming $\phi(.)$ as $\sqrt{.}$ we get,\\
\begin{equation}
\begin{aligned}
f_{FB}(A) &= \sum_{i \in U} \gamma_i\psi(\sum_{a \in A} \sigma(U_i^a \theta_i))\\
\triangledown_{\gamma_i}f_{FB} &= \psi_{i} (m_i(A))\\
\triangledown_\theta f_{FB} &= \sum_{i \in U} \gamma_i\frac{\sum_{a \in A} \sigma(U_i^a \theta_i) (1 - \sigma(U_i^a \theta_i)) U_i^a}{2\sqrt{\sum_{a \in A} \sigma(U_i^a \theta_i)}}
\end{aligned}
\end{equation}

\subsubsection{Query-focused summarization}
Consider an entry $sq_{ij}$ from the similarity matrix $S_Q$ which is computed from the off-the-shelf/raw feature representation of query $q_i$ and DSN feature representation of data point $X_j=\sigma(\theta^T \cdot U_j)$. Assume that $X_j$ and $q_i$ are L2 normalized. Then, the cosine similarity between them is simply their dot product: $s_{ij} = q_i^T \cdot X_i$. We can rewrite $sq_{ij} =  q_j^T \cdot \sigma(\theta^T \cdot U_i)^T$, where $\sigma$ is the sigmoid function. The subgradient of the similarity matrix $\triangledown_\theta sq_{ij}$ can be computed as follows:

\begin{align}
    \triangledown_\theta sq_{ij} = q_i^T \sigma(\theta^T U_j )(1- \sigma(\theta^T U_j ))U_j
\end{align}
We compute the gradients wrt $\lambda$ and $\theta$ as follows:\\
\textbf{GCMI:} From \ref{eq:GCMI} we get,
\begin{align}
    \triangledown_\theta I_f (A;Q) = 2\sum_{i \in A} \sum_{j \in Q}  \triangledown_\theta sq_{ij}
\end{align}

\noindent\textbf{FL1MI:} From \ref{eq:FL1MI} we get:
\begin{equation}
\begin{aligned}
    \triangledown_\lambda I_f (A;Q) &= \sum_{i \in V} \max_{j \in Q} sq_{ij}*1_{\lambda sq_{ij_Q} \leq s_{ij_A})}\\
    \triangledown_\theta I_f (A;Q) &= \sum_{i \in V} \triangledown_\theta s_{ij_A} * 1_{s_{ij_A} < \lambda sq_{ij_Q}} + \lambda \triangledown_\theta sq_{ij_Q} * 1_{\lambda sq_{ij_Q} \leq s_{ij_A}}
\end{aligned}   
\end{equation}
where, $j_A = \argmax{j \in A} s_{ij}$ and $j_Q = \argmax{j \in Q} sq_{ij}$

\noindent\textbf{FL2MI:} From \ref{eq:FL2MI} we get,
\begin{equation}
\begin{aligned}
    \triangledown_\lambda I_f (A;Q) &= \sum_{i \in A} \max_{j \in Q} sq_{ij}\\
    \triangledown_\theta I_f (A;Q) &= \sum_{i \in Q} \triangledown_\theta  sq_{ij_A} + \lambda \sum_{i \in Y} \triangledown_\theta  sq_{ij_Q}
\end{aligned}   
\end{equation}
 where, $j_A = \argmax{j \in A} sq_{ij}$ and $j_Q = \argmax{j \in Q} sq_{ij}$

\begin{algorithm}
\caption{DSN Learning}
\label{alg:dsnlearn}
\begin{algorithmic}[1]
\STATE \textbf{Input:} $N$ Training examples $\{X^n\}_{n=1}^N$ and their ground truth summaries $\{Y^{(n)}\}_{n=1}^N$, M submodular functions $\{f_i\}_{i=1}^M$\\
\STATE \textbf{Initialize} $\Theta = (w_1 \dots w_M, \lambda_1 \dots \lambda_M, \theta)$
\WHILE{Not Converged}
\STATE{For fixed $( \lambda_1 \dots \lambda_M, \theta)$ update $(w_1 \dots w_M)$ by computing subgradients using \ref{eq2}}
\STATE{For fixed $(w_1 \dots w_M)$ update  $( \lambda_1 \dots \lambda_M, \theta)$ by computing subgradients using \ref{eq3} and \ref{eq4}}
\ENDWHILE\\
\textbf{Output: } Optimal DSN parameters $\Theta$
\end{algorithmic}
\end{algorithm}

\section{Empirical Experiments for Learning using DSN}

In this section, we evaluate the performance of our method, which we refer to as Deep Submodular Networks (DSN), and compare it with 1) submodular mixture model \cite{tschiatschek2014learning}, 2) Deep Submodular Functions (DSF)~\cite{dolhansky2016deep}, Differentiable Submodular Maximization (DSM)~\cite{tschiatschek2018differentiable}, and individual submodular functions. We use the data set of \cite{tschiatschek2014learning} (and its extension~\cite{kaushal2020unified}) consisting of 14 distinct real-world image collections with 100 images per collection. The image summarization problem is to select a summary set of 10 images that best represent the collection. We train the DSN on each of the 14 collections by using a leave-one-out-cross-validation scheme \cite{hastie2009elements} on the remaining 13 collections. We use the max-margin approach for training where the loss function is $l(A) = 1-r(A)$ and $r(A)$ is V-ROUGE (as defined in section 2.2 of \cite{tschiatschek2014learning}). For evaluation, we use a normalized version of V-ROUGE where the average normalized V-ROUGE score of random summaries is 0, and that of human summaries is 1. 

\subsection{Generic Summarization}
We used a DSN of the form $F(A, x^{(n)}, w, \theta, \lambda) = \sum_{i=1}^M w_i f_i(A, x^{(n)}, \theta, \lambda_i)$. For the parameterized feature embedder, we use a hidden layer of 512 units of sigmoid activation whose weights are initialized using Xavier initialization \cite{glorot2010understanding}. The similarity kernel $S$ is computed using cosine similarity. For our experiments, we initialize the DSN with either a single fucntion or a mixture of four submodular functions - namely, graph-cut (\ref{eq:GraphCut}), saturate coverage (\ref{eq:SatCov}), facility location (\ref{eq:FL}) or feature based (\ref{eq:FB}). The mixture weights are randomly initialized and the DSN was trained end-to-end using Adam and a decaying learning rate.

\begin{figure}[h]
    \centering
    \includegraphics[width=0.5\textwidth]{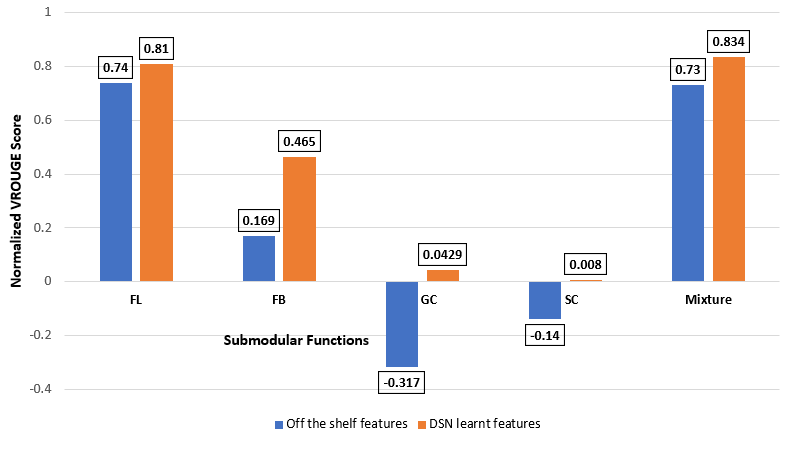}
    \caption{Performance of DSN for Generic Summarization on Test data.}
    \label{fig:generic_dsn}
\end{figure}

\begin{figure}[h]
    \centering
    \includegraphics[width=0.9\textwidth]{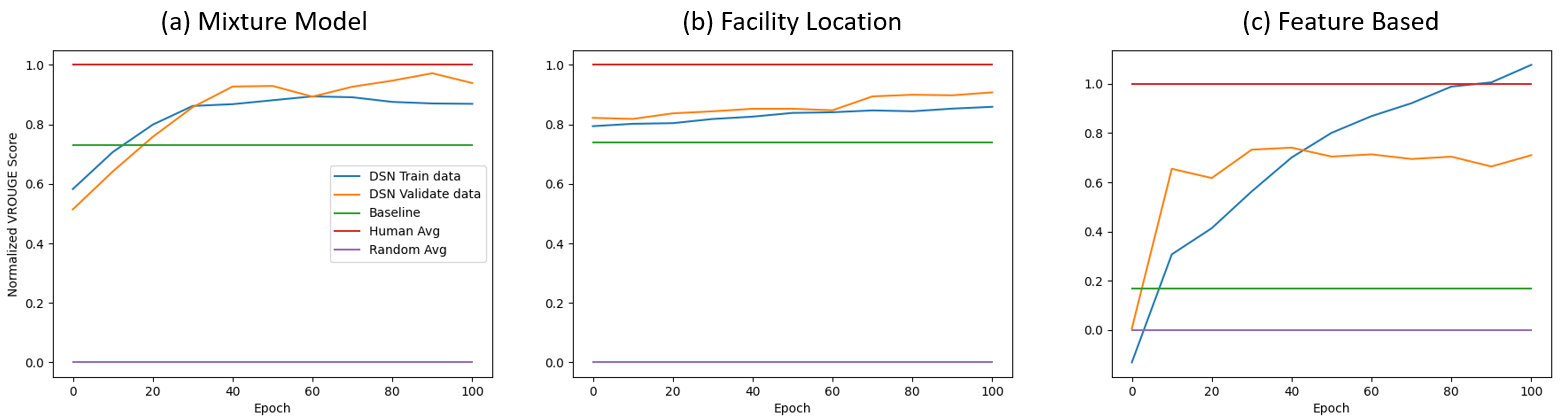}
    \caption{Learning and generalization of DSN. Performance of - Left: Mixture Model with 4 components, Middle: facility location, Right: feature based function}
    \label{fig:dsn_plot}
\end{figure}

In the first set of experiments, we represent each image using features extracted from the pool5 layer of GoogleNet \cite{szegedy2015going} pre-trained on ImageNet \cite{deng2009imagenet}. We compare submodular functions instantiated using features from the above pre-trained DNN with submodular functions instantiated using learned features from a DSN. Figure \ref{fig:generic_dsn} shows that DSN learned features outperform the pre-trained DNN features for all submodular functions. This performance shows that DSN can indeed learn the features and individual parameters of the four submodular functions (FL, FB, GC and SC). Next, we repeat this for the mixture. We compare the mixture of DSNs learnt in an end to end manner, with a simple mixture model over the four submodular functions. Again, we see that Mixture-DSNs significantly outperforms just the mixture. Figure \ref{fig:dsn_plot}, we illustrate the learning of DSNs over epochs (comparing both the training and validation set performance) in both the mixture case (Figure \ref{fig:dsn_plot} (a)) or individual functions FL or FB (Figure \ref{fig:dsn_plot} (b) and (c)). 

\begin{figure}[h]
    \centering
    \includegraphics[width=0.45\textwidth]{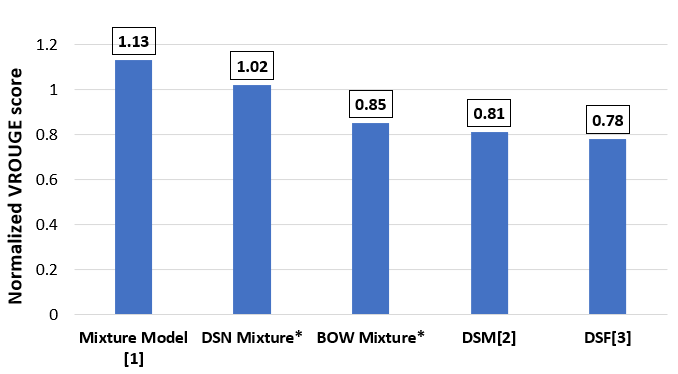}
    ~
        \includegraphics[width=0.45\textwidth]{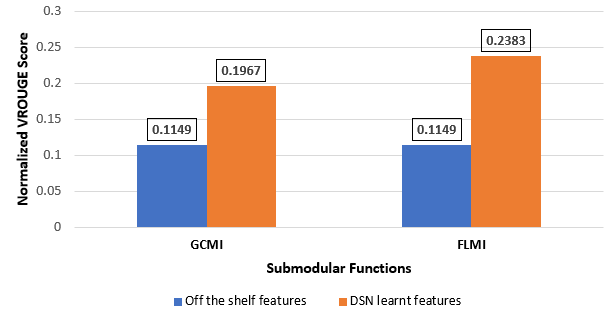}
    \caption{\textbf{(Left Figure):} Comparison of DSN with other methods. In all methods,  submodular functions were instantiated using visual bag-of-words feature representation. *The DSN Mixture and BOW Mixture using the same components. \textbf{(Right Figure):} Performance of DSN for Query-focused Summarization on Test data.}
    \label{fig:compare_dsn}
\end{figure}

In the second set of experiments, each image is represented using visual-word features (as described in Section 5 of \cite{tschiatschek2014learning}). We do this to compare with the mixture model of \cite{tschiatschek2014learning}, which uses the BOW features instantiated with 594 components of submodular functions. In contrast, we define our DSN only with four components. The results are shown in the left figure of Figure~\ref{fig:compare_dsn}. We achieve a VROUGE score of 1.02 which is close to the 1.13 achieved using 594 components (which is also significantly more expensive for both inference and learning). We also compare against a simple mixture model using the same four components, which achieves a VROUGE score of 0.85 (BOW-Mixture in the figure). We additionally compare against DSFs~\cite{dolhansky2016deep} and DSMs~\cite{tschiatschek2018differentiable}, both instantiated with BOW features which achieve 0.78 and 0.81 respectively. 




\subsection{Query-focused summarization}

We used a DSN of the form $F(A, Q, x^{(n)}, w, \theta, \lambda) = \sum_{i=1}^M w_i I_{f_i}(A, Q, x^{(n)}, \theta, \lambda_i)$. For the parameterized feature embedder, we use a hidden layer of 959 units of sigmoid activation whose weights are initialized using Xavier initialization \cite{glorot2010understanding}. The similarity kernels $S$ and $S_Q$ are computed using cosine similarity. For these set of experiments, we initialize the DSN with either GCMI (\ref{eq:GCMI}) or FL2MI (\ref{eq:FL2MI}). The mixture weights are randomly initialized, and the DSN was trained end-to-end using Adam and a decaying learning rate. We used the query-image summarization dataset~\cite{kaushal2020unified,tschiatschek2014learning}.

We instantiate the DSN's mixture components by using a concatenated probabilistic feature vector obtained from the YOLOv3 model \cite{redmon2018yolov3} pretrained on OpenImages and a scenes model \cite{zhou2014learning} pretrained on Places365 dataset. Figure \ref{fig:compare_dsn} (right figure) compares SMI functions instantiated with DSN learned features to SMI functions instantiated with the pre-trained features, and we see that the DSN functions significantly outperform the functions without joint feature learning.

\section{Conclusions}
In this paper, we presented Deep Submodular Networks, which enables us to jointly learn features along with the submodular function parameters. We study the parameter learning problem for this network using the max-margin learning framework, which enables end-end learning of both the submodular network parameters (weights and internal parameters), and the joint features. We demonstrate the utility of DSNs on both generic and query focused image-collection summarization, and show significant improvement over the state-of-the-art. In particular, we show that DSNs outperform simple mixture models using off the shelf features. Secondly, we also show that just using four submodular functions in a DSN with end-to-end learning performs comparably to the state-of-the-art mixture model with a hand-crafted set of 594 components~\cite{tschiatschek2014learning} and outperforms other methods \cite{tschiatschek2018differentiable, dolhansky2016deep} for image collection summarization.

\bibliographystyle{unsrt}  

\bibliography{main.bib}





\end{document}